# Evaluation of Human-Robot Collaboration Models for Fluent Operations in Industrial Tasks

Lior Sayfeld, Ygal Peretz, Roy Someshwar, Yael Edan
Dept. of Industrial Engineering and Management,
Ben-Gurion University of the Negev, Beersheba, Israel 84105
Email - sayfeld@post.bgu.ac.il, igalpe@post.bgu.ac.il, royso@post.bgu.ac.il, yael@bgu.ac.il

*Abstract* — In this study we evaluated human-robot collaboration models in an integrated human-robot operational system. An integrated work cell which includes a robotic arm working collaboratively with a human worker was specially designed for executing a real-time assembly task. Eighty industrial engineering students aged 22-27 participated in experiments in which timing and sensor based models were compared to an adaptive model developed within this framework. Performance measures included total assembly time and total idle time. The results showed conclusively that the adaptive system improved the examined parameters and provided an improvement of 7% in total assembly time and 60% in total idle time when compared to timing and sensory based models.

## I. INTRODUCTION

The de facto understanding in industrial robotics is that robots are supposed to be enclosed in cages to avoid the slightest possibility of any intersection between human and robot working space. Current R&D trends aim to integrate the human worker into the robot workspace to take advantage of both systems thereby increasing profitability and efficiency [1], [2] [3]. Cooperation between two humans is inherently intuitive [4]; however collaboration between a human and robot introduces challenges [2], [5] that must be investigated in order to provide an efficient and user-friendly system. In a collaborative system, both participating partners must know how to adapt themselves to each other to perform effectively the action in the right time and space. One of the major challenges is how to provide the robot with the capability to communicate and adapt itself to the human in a way that the work/job/action would be performed optimally.

Most collaborative algorithms to date are primarily based on the "Master-Slave" principle in which the human worker operates the robot in a direct way or programming it a-priori off-line. This allows the robot to do only static and repetitive actions. To ensure worker safety, the work environments of robots and human are completely separated in time and place. "The collaboration between human flexibility and robot efficiency can essentially reduce production costs and increase/enhance production rate and efficiency"[6]. In order to achieve this kind of collaboration there is a need for safe and adaptive robots that will be able to learn and understand the worker in front of them in order to help him execute his task/job. A right combination of human's and robot's strengths for performing a collaborated task enables a more accurate and efficient system. Current human robot collaboration systems are based mainly on the principal of "Stop & Go" and "Turn Taking". That is, the robot performs a specific action, finishes it and only afterwards the human worker starts his actions and vice versa. In order that the robot will operate in a more 'friendly' and 'humanely' way during the collaboration, the interaction between them should be continuous and fluent like the interaction between two people [7]. In previous works [8] [9], we developed three models for human–robot collaboration – Timing, Sensor and Adaptive. In the timing based model [10] the robot repeats actions defined in advance at specific time intervals that are defined by the operator. In the sensor based model the robot performs the action only when signals are received from the sensors. In the adaptive model the robot matches its working pace in real time to the human action (not necessarily a physical action but any interaction between human and robot).

Analysis of these H-R collaboration models were carried out previously using simulation tools [10] and analytical tools [9]. These analyses have shown how influencing parameters affect the collaboration process and how each of the models can turn out to be the best model of collaboration under different circumstances depending upon the scenario. The aim of the current study was to perform an experimental analysis of the three models with human subjects simulating a real-life scenario of a human and a robot working together in a factory floor. The models were implemented in an operational system and their performances were evaluated for a slow paced human-robot collaborative task.

## II. METHODOLOGY

### A. System

The system (fig.1) includes a 5 DOF revolute robotic arm (Scorbot ER4U) and a computerized operating system in which the three algorithms were implemented. The human and the robot collaborate for executing a real-time assembly task (fig.3) which is to build a tower from two kinds of LEGO cubes – A and B (fig.3). The robotic arm is responsible for delivering the bigger cube (A) to the human in every assembly cycle. The

This research was partially supported by the Helmsley Charitable Trust through the Agricultural, Biological and Cognitive Robotics Initiative and by the Rabbi W. Gunther Plaut Chair in Manufacturing Engineering, both at Ben-Gurion University of the Negev.

human operation is divided into two tasks: 1) fetching the cube (A) from the robotic arm and connecting it on top of the preceding cube and, 2) Picking small cubes (B) from a designated container and assembling them layer by layer around the periphery of the tower. The total task in each cycle consisted of building 4 layers, which in turn means, fetching 5 bigger cubes (A) from the robot in 5 different cycles and fetching 20 cubes of size (B) from a container, but only one at a time and only after finishing each time the repetitive task of putting cube B around the periphery of cube (A). The total cycle time was 317 seconds (average).

A secondary task for the robot was included in the system with the intention to simulate a scenario of a multi-tasking robot which aims to maximize its efficiency and resources by utilizing the available waiting time in between two cycles for another task that may be completely secondary in nature or one that augments the overall assembly cycle time. In this study, we added a secondary task in which the robot kept refilling the cube (A) buffer between two successive cycles.

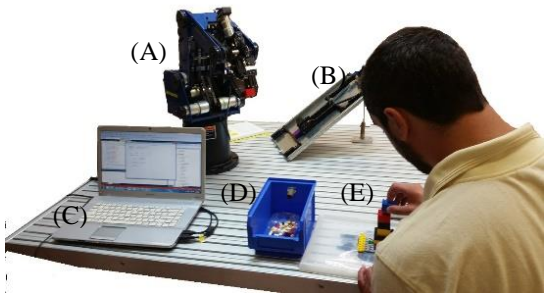

Figure 1: (A) The robot, (B) The secondary task buffer, (C) Operating system, (D) Cubes (B) container, (E) Assembly task

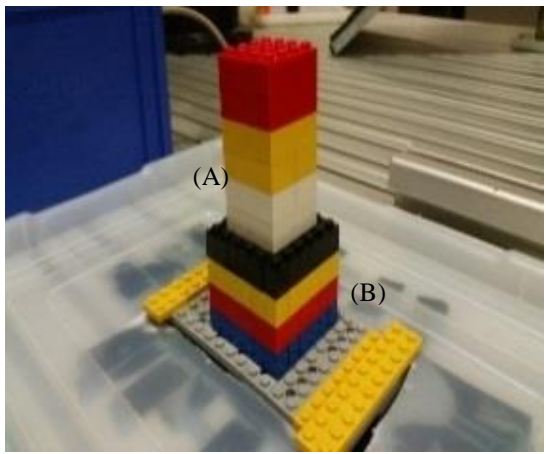

Figure 2: Cube (A) and cube (B) of the assembly task

*Algorithms*
Three algorithms, detailed in section III, were implemented using C++ to control the robot: Timing, Sensor and Adaptive.

*Performance measures*
The performance measures selected were total assembly time and total idle time (sum of human and robot idle time). Both measures were examined with 95% significance level ($\mu$ refers to the mean of each indicator).

$$H_0: \mu_{timing} - \mu_{adaptive} > 0$$
$$H_0: \mu_{sensor} - \mu_{adaptive} > 0$$

### B. The Experiment

Experiments were performed at the Integrated Manufacturing Technology (IMT) laboratory of Ben-Gurion University of the Negev. Eighty undergraduate students from the Dept. of Industrial Engineering (34 female, 46 male) aged 22-27 participated in the experiments which included a collaborative work with the robot followed by a questionnaire. The questionnaire included subjective measures which is part of a different study and is not the focus of this paper. The experiment process itself took about 30 minutes in which each subject had to build a tower collaboratively with the robot. The human and the robot repeated this collaborative task three times denoted in this experiment as three rounds.

Each time a different algorithm was selected randomly and executed. Between each round the participants were given two minutes of rest. All data collection was automated through the integrated computerized system.

### C. Analysis

Analyses were conducted in SPSS software using randomized block design in order to remove differences among the experimental subjects within a particular model.

### III. ALGORITHMS

### A. Timing

The *timing* algorithm uses a fixed cycle time with time intervals set at 70 sec. The calculation of this value was based on the average of a set of 10 subjects from the preliminary experiment in which human action times were measured.

### B. Sensor based Algorithm

The *sensor based* algorithm is based on signals it receives from an infra-red (IR) proximity sensor placed at the work-space of the human. This sensor sends a signal for every time the human picks up a small cube (B). The action-triggering signal is the one that is sent when the human picks up the 13$^{th}$ cube. The robot is informed that the preceding act of the collaborating human is about to end in a designated amount of time. As a result, the robot finishes the immediate secondary task at hand and immediately initiates its primary task of handing over the bigger cube (A) to the human.

### C. Adaptive Algorithm

The basis of this algorithm lies in a formula that predicts the assembly time of the following cycle. The input for this formula is the time required by the human to place each individual small cube (B) and the predicted output is the cycle time of the successive cycle of the assembling stage. The total work is divided between the human and the robot and both of them work simultaneously. To adapt the robot's operations to human speed, the algorithm calculates the assembly time of the previous cycle by calculating the time difference between them; this time is then sent to the prediction formula and the predicted total assembly time is received; given this time the robot can plan its secondary work and perform it until the person is ready for the next handover.

The prediction formula includes parameters that vary along the task due to the uneven pace of the human and due to rate changes caused by unplanned situations. Three main parameters were weighted in the formula: average assembly time of the general population, the current person's average assembly time and a moving average of the assembly time of the last three pieces that a person assembled. The weights were designed so as to allow the algorithm to react quickly to changes in the work rate (by providing a larger weight to the three last parts of the prediction formula) and "dampen" the predicted change. Below are details of the prediction formula:

$$F = (20-n)(\alpha * D_T + \beta * D_S + \gamma * (\delta * P_n + \varepsilon * P_{n-1} + \theta * P_{n-2}))$$

($n$-number of cubes (A) assembled, $D_T$-general mean of population for assembly of one cube, $D_S$- mean of the current human for assembly of one cube, $P_n$-mean of assembly time for the-n cube)

The weights of the parameters used in this formula are not fixed and vary on-line according to the work processes. The values of the weights are determined by a control process performed within this prediction formula. This control process classifies the size of the error in the formula by comparing the forecasted value to the real assembly time measured by the system and tries to "repair" it by changing the weights.

IV. RESULTS

The effect of the block and the main factor was significant (sig 0.00) for the assembly time measure, therefore we used one way model with Tukey as a Post-Hok test. For the Idle time measure the block shows no effect (sig 0.192) unlike the main factor which was significant (sig 0.00), therefore we used the one way model without block with Tukey as a Post-Hok test. The conducted tests executed the comparison between the means of each alternative.

A. Total assembly time

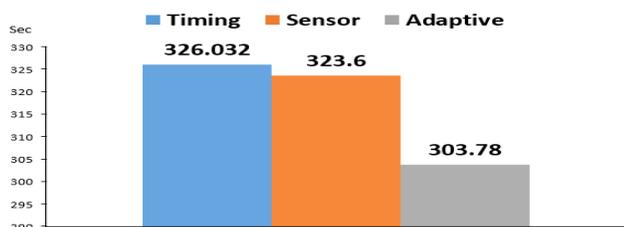

Fig: The total assembly time taken in Timing, Sensor and Adaptive Model

The average total time of the adaptive algorithm was significantly (sig 0.00) lower than the sensor based model by 7% and by 14% from the timing based model. There was no significant (sig 0.103) difference between the timing and sensor based model.

B. Total idle time

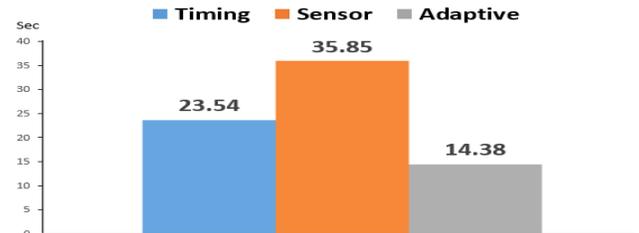

Fig: The total idle time (including human and robot) for Timing, Sensor and Adaptive Model

The total idle time of the adaptive algorithm was significantly (sig 0.00) lower than the sensor based model by 60% and lower than the timing based model by 39%. The timing based model was significantly (sig 0.00) lower than the sensor based model by 35%.

V. DISCUSSION AND CONCLUSION

An integrated human-robot collaborative system was developed and implemented in an experimental framework which simulated an industrial assembly task. Results showed conclusively that in the adaptive system both the total assembly time and idle times are reduced (by 7% and 60% respectively).

This work indicates that improvement in the efficiency and productivity of the production line can be achieved in a collaborative system. Ongoing research is aimed at developing advanced adaptive algorithms to further improve performance. Evaluation will be conducted for a variety of tasks. Additionally, the influence of psychological and physiological aspects of the person who works with the robot will be analyzed to provide a deeper understanding on the influencing parameters.